RESEARCH ARTICLE OPEN ACCESS

# Adaptive Collision Sensitivity for Efficient and Safe Human–Robot Collaboration

Lukas Rustler | Matej Misar | Matej Hoffmann

Department of Cybernetics, Faculty of Electrical Engineering, Czech Technical University in Prague, Prague, Czech Republic

**Correspondence:** Matej Hoffmann (matej.hoffmann@fel.cvut.cz)

## ABSTRACT

What is considered safe for a robot operator during physical human–robot collaboration (HRC) is specified in corresponding HRC standards (e.g., ISO 10218-2:2025). The regime that allows collisions between the moving robot and the operator, called Power and Force Limiting (PFL), restricts the permissible contact forces. Using the same fixed contact thresholds on the entire robot surface results unnecessary productivity losses, as the robot needs to stop even when impact forces are within limits. Here we present a framework that decides whether the robot should interrupt motion based on estimated collision force computed individually for different parts of the robot body and dynamically on the fly, based on the Effective Mass (EM) of each robot link. We performed experiments on a simulated and real collaborative robot arm (UR10e) with sensitive skin (AIRSKIN) for collision detection and isolation. To demonstrate the generality of our method, we added experiments on simulated KUKA LBR iiwa robot, where collision sensing draws on joint torque sensing. On a mock pick-and-place scenario with both transient and quasi-static collisions, we show an increase in productivity over 45% from using the standard approach. The method is applicable to any robot for which the EF can be calculated.

## 1 | Introduction

Robots are leaving safety fences and starting to share workspace or even living space with humans. As they leave controlled environments and enter domains that are much less structured, they need to dynamically adapt to unpredictable interactions and ensure safety at every moment. The last decade has seen rapid development in the field of physical human–robot interaction (pHRI) (see [1–4] for surveys), alongside a growing market for collaborative robots and the introduction of new safety standards [5–7]. According to ISO 10218-2:2025 [6] (formerly ISO/TS 15066 [7]), there are two main ways to satisfy the safety requirements when a human physically collaborates with a robot: (i) Power and Force Limiting (PFL) and (ii) Speed and Separation Monitoring (SSM). For PFL, physical contacts with a moving robot are allowed, but the forces, pressures, and energy absorbed during a collision must be within part-specific limits of the human body. This translates into lightweight structure, soft padding, and no pinch points on the robot side, in combination with robot interaction control methods for the post-impact phase: collision detection and response relying on motor load measurements, force/torque, or joint torque sensing (e.g. [8]). However, the speed and payload of robots that comply with this safety requirement is limited.

According to Haddadin et al. [9], a collision needs to be 1) detected, 2) isolated (which robot link), 3) identified (estimating the external torque or force), 4) classified and 5) resolved by triggering an appropriate robot reaction—see Figure 1. The accuracy or even the possibility of performing some of these steps depends on the type of contact (single or multiple), the contact location, and in particular on the sensory equipment of the robot. Electronic skins covering large parts of the robot surface [10–13] facilitate detection and isolation. For identification, the external torque or force action on the robot needs to be estimated, for which calibrated sensitive skin or torque sensors and a model of the robot dynamics need to be available. However, such estimations are never perfectly accurate. Moreover, the assumption from Newton's third law that the same force acts on the human body may not hold perfectly. In addition, to comply with the standards, the contact surface area and the affected human body part would have to be known. In practice, collaborative robot applications are configured too conservatively, that is, stopping the robot as soon as a collision is detected without any further analysis. Frequent stops of the robot then negatively impact productivity—and it may also impact the health of the robot components.

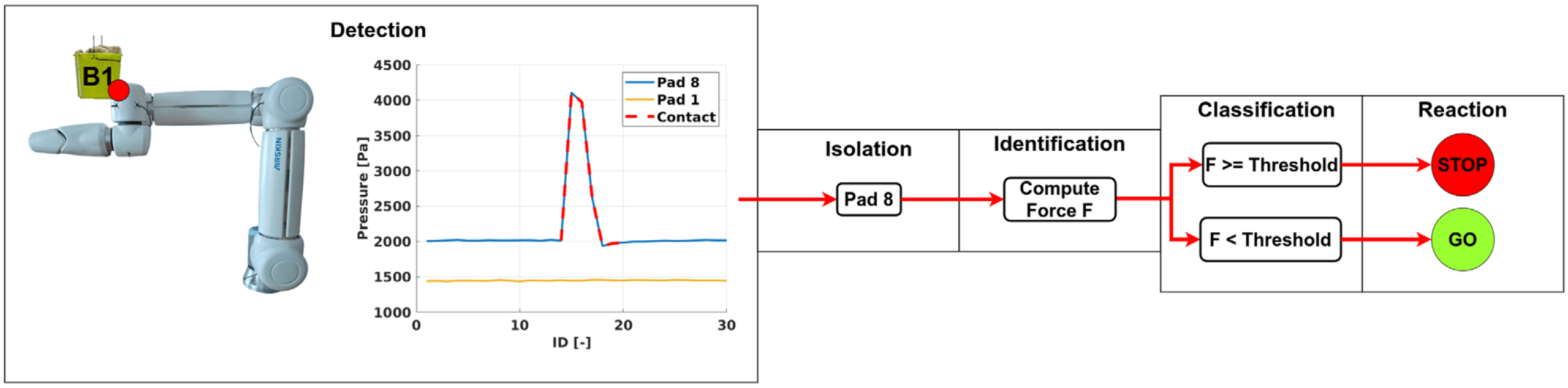

**FIGURE 1** | Schematic overview of collision handling pipeline adapted from [9]. In Power and Force Limiting (PFL) mode, we care about the collision and post-collision phases. The collision needs to be first detected. Then further analysis can be performed to determine the position of contact (isolation) or the forces exerted during the contact (identification). Based on these, the type of contact is classified and proper reaction is executed. In our case, we classify whether the contact is higher or lower than a given threshold and the robot is instructed to either continue the task (force under the threshold) or stop (force over the threshold).

In this work, we present a new method that aims to maximize productivity in the PFL regime, as illustrated in Figure 1. Our approach detects and isolates collisions using sensors (e.g., robotic skin or joint-torque sensors), identifies the Estimated Collision Force ($F_{est}$) via a simplified dynamic model, and triggers an appropriate real-time robot reaction. We tested the method on two robots, and the collision is detected and isolated using a pressure-sensitive[1] electronic skin (AIRSKIN) covering the entire robot body of the UR10e robot (both real and simulated), or using joint-torque sensors of the KUKA LBR iiwa 7 robotic arm (only simulated). We strictly utilize physical sensors to detect contact events and isolate the specific link or skin pad involved, rather than using raw sensor readings directly to estimate $F_{est}$. The key innovation here is that we estimate $F_{est}$ in real time (25 Hz when using all 11 skin pads on our UR10e robot) from the robot velocity and Effective Mass (EF) and then use that estimate to decide how to react to the collision. If $F_{est}$ is less than a threshold, the robot will react by continuing to perform the task. If $F_{est}$ is greater than a threshold, the robot will stop.

To our knowledge, this is the first study that adaptively sets collision thresholds of a collaborative robot that brings safety (force limits according to International Organization for Standardization [6]) and productivity under one hood. Solutions for use in industry such as AIRSKIN normally employ the same threshold in all skin pads and use a highly sensitive setting, resulting in too frequent task interruptions. In research articles employing large-area skins, the same thresholds are also typically employed [14].

While the preliminary conceptual idea of using EF to modulate collision reactions was first introduced in our conference work [15], that initial study was primarily simulation-based, featured only preliminary real-world tests and was not strictly formulated around standardized safety limits. This article consolidates that preliminary concept into a systematically validated framework. First, the collision handling architecture has been fundamentally redesigned to guarantee safety. Previously, a collision was detected only if the raw pressure measured inside the skin exceeded a predefined threshold, which posed the risk of missing low-pressure but potentially unsafe impacts. In our current implementation, every contact event is detected immediately, and the post-collision decision (stop versus continue) is safely governed solely by $F_{est}$. Second, we provide a vastly expanded, systematic experimental validation by: (i) introducing quasi-static (clamping) impact evaluations alongside transient contacts; (ii) demonstrating the general applicability of the framework by adding a second robot (KUKA LBR iiwa 7) that draws on joint-torque sensing rather than electronic skin; and (iii) conducting rigorous evaluations across multiple Cartesian velocities to map the exact trade-offs between productivity gains and safety compliance.

## 2 | Method

The core of this work involves systematically evaluating simulated and real-world impacts within an experimental setup designed to emulate human–robot collaboration alongside static and dynamic obstacles. This section details the theoretical foundations, implementation architecture, and robotic platforms utilized.

### 2.1 | Task and Environment

The experimental environment (Figure 2) consists of three buckets (B1, B2, B3) suspended from the ceiling by ropes. The buckets are filled with small stones. The buckets can freely move in space, thus simulating the impact of the robot with the human arm. The impacts with buckets are transient, that is, dynamic impacts in which a human body part is impacted by a moving robot and the human can recoil or retract [6]. The setup also includes the *CBSF-75-Basic* certified measuring device for the validation of collaborative tasks in the range of 20–500 N. This device is firmly attached to the table and is used to simulate a clamping scenario. This impact is quasi-static, that is, the impact of a human body part with the robot, when the robot clamps or crushes the human for a prolonged period of time [6]. We had to put a piece of foam on top of the device (see Figure 2) to avoid the robot's internal safety controller from interrupting the experiments. The UR10e robot has an internal safety controller that can interrupt the task when the current in the motors is above a given threshold or when the robot deviates from its trajectory. There are only pre-defined safety settings available, and unfortunately, even with the least restrictive one, the robot was sometimes unable to stop on our command fast enough before triggering the internal stop. The foam added more time for the robot to stop properly based on a command from our pipeline. We are aware that the padding changes the Measured Collision Force ($F_{mes}$). However, the modification was necessary to run the experiments. The foam used has a thickness of 30 mm and a hardness of 3.6 kPa, and based on our test the measured $F_{mes}$ are reduced from the Real Impact Force ($F_{real}$) by approximately 20%.

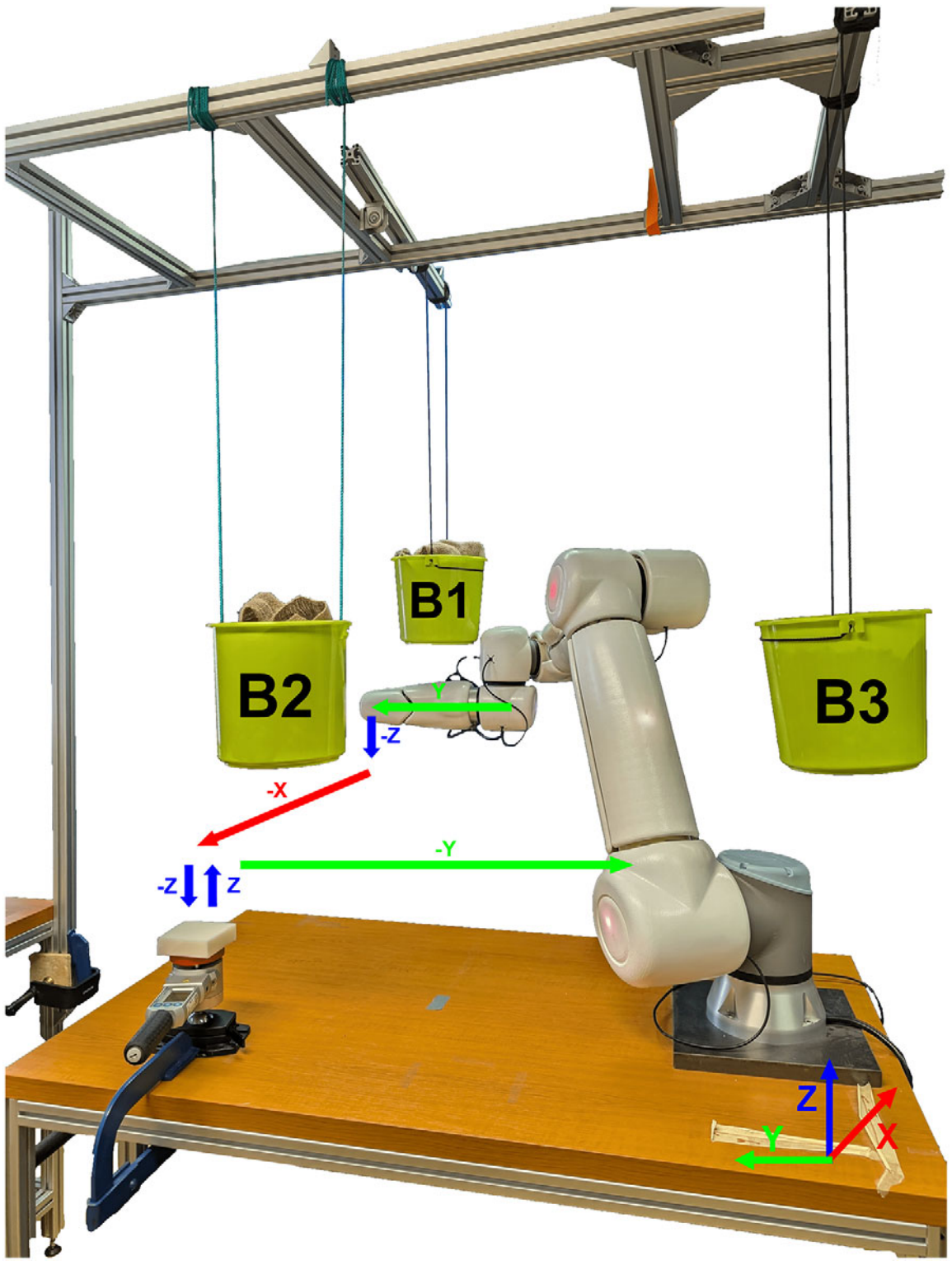

**FIGURE 2** | The environment used in the experiments. The hanging buckets simulate transient contacts. The impact measuring device fixed on a table simulates quasi-static contact (clamping scenario). The arrows depict the directions of movement during a mock pick-and-place scenario used in this work.

To be able to evaluate the effect of different collision threshold settings on performance, we define a mock pick-and-place task—see the video at https://youtu.be/UyOa--6Jpj4. The direction of the movements is depicted in Figure 2. The order of movements starts from the top green line (movement in the $y$-axis of the robot) and collision with the first bucket B1, continues with slight downward movement (negative $z$-axis) and movement in the robot's $x$-axis. Then it continues with impact to B2. The robot then goes again downward to collide with the measuring device, and returns up after the collision. Finally, the robot moves in its negative $y$-axis and collides with bucket B3.

## 2.2 | Robot, Skin, and Software

We have used two robots in this work. The robots are depicted in Figure 4 and are detailed in the following section.

### 2.2.1 | *Universal Robots UR10e with AIRSKIN*

This robot is a 6-Degree of Freedom (DoF) robotic manipulator, and the arm itself is designed for collaborative purposes. Our robot is further equipped with robotic skin AIRSKIN.[2] Using a special AIRSKIN maintenance controller, the pressure values from individual skin parts (also called pads; 11 of them placed on our robot) can be retrieved at approximately 30 Hz. Unlike in the standard AIRSKIN deployment in the industry where any detected collision stops the robot, here we use the pressure sensor for collision detection and isolation (which robot link collided) and choose to stop or continue based on a force $F_{est}$ computed by our algorithm. A real and simulated version of the robot was used.

### 2.2.2 | *KUKA LBR iiwa 7*

The second robot used is also collaborative. It is equipped with joint torque sensors in each of its joints and can estimate external torques, allowing detection and isolation of collisions. A simulated version of the robot was used.

### 2.2.3 | *Control, Communication, and Environment*

Both robots are integrated and controlled using the Robot Operating System (ROS) middleware. The artificial skin is also integrated into ROS. The task performed by the robots (see Section 2.1) is defined in the Cartesian space, but we utilize control in the joint velocity space because it gives us more control over the movements and contact reactions.

During the experiments, not all skin pads or robot links were in contact with the obstacles—the ones used are colored in Figure 4. The choice of robot links that will collide with the environment comes from the desired robot trajectory and obstacle placement and implies no loss of generality of our method.

### 2.2.4 | *Simulation*

We used simulators that we developed (available at https://rustlluk.github.io/adaptive-sensitivity). For both robots, the underlying physics simulation is achieved with PyBullet [16]. The simulator also contains the buckets and force-measuring device with realistic physical behavior allowing us to do a higher number of experiments than in the real world.

## 2.3 | Collision Handling Strategies

We employ three different strategies to handle collision events: (i) Factory (FA); (ii) Fixed Mass (FM); (iii) Adaptive Mass (AM). The first two are taken as baselines and the last one is our contribution. These can be further divided into two categories: static and dynamic collision handling.

### 2.3.1 | *Static*

The representative of this group is FA, which simulates the common ("industrial") strategy used with collaborative robots in industry: Whenever a contact is detected, the robot is stopped immediately. No further analysis of the contact (e.g., collision isolation or identification) is used. This serves as a baseline to gauge how much productivity can be gained when a better collision reaction strategy is used.

### 2.3.2 | *Dynamic*

The two strategies described below (FM and AM) are dynamic, that is, the robot reacts to a collision by stopping or continuing based on the Estimated Collision Force ($F_{\text{est}}$) computed on the fly using the following equation:

$$F_{\text{est}} = \frac{v\sqrt{k}}{\sqrt{m_{\text{R}}^{-1} + m_{\text{H}}^{-1}}} \tag{1}$$

where $k$ is the spring constant of the impacted human body part, $m_{\text{R}}$ is the mass of the robot, $m_{\text{H}}$ is the mass of the impacted human body part, and $v$ is the relative Cartesian velocity of the human and robot (consistently to the ISO we assume static humans, and thus this simplifies to the velocity of the robot link in contact).

We evaluate an instance of this equation for each robot link on the run, inputting the instantaneous velocity $v$ of every robot link. If a collision of a particular robot link is detected, the equation is used to estimate the impact force $F_{\text{est}}$.

The two dynamic methods to compute collision thresholds differ in the way they determine the mass of the robot $m_{\text{R}}$.

For FM, the mass $m_{\text{R}}$ is defined as Half of Moving Mass (HMM) of the robot (as in ISO 10218-2:2025). However, that mass would be constant for each collision point. We were seeking a baseline that reflects the collision point on the robot and chose a naive approximation as half of the sum of the masses of links from the first moving link up to the colliding link (including). This provides an adaptive and harder to beat baseline—the masses are lower than the HMM of the whole robot (see *HMM Full* in Figure 3). Note, that this method is not strictly conservative.

In the case of AM, $m_{\text{R}}$ is the real Effective Mass (EF) of the robot. EF is a property of the robot (of each point of the robot), which can be computed using the dynamics of the robot [17]. EF is always computed in a given direction—a normalized 3D directional vector. Using this direction $\mathbf{u}$, one can compute the EF $m_{\mathbf{u}}$ as

$$m_{\mathbf{u}} = \frac{1}{\mathbf{u}^T \boldsymbol{\Lambda}_v(\mathbf{q})^{-1}\mathbf{u}} \tag{2}$$

where $\boldsymbol{\Lambda}_v(\mathbf{q})^{-1}$ is the upper $3 \times 3$ matrix of the Cartesian kinetic energy matrix $\boldsymbol{\Lambda}(\mathbf{q})^{-1}$, which can be calculated as

$$\boldsymbol{\Lambda}(\mathbf{q})^{-1} = \mathbf{J}(\mathbf{q})\mathbf{M}(\mathbf{q})^{-1}\mathbf{J}(\mathbf{q})^{\text{T}} \tag{3}$$

where $\mathbf{M(q)}$ is the joint space inertia matrix and $\mathbf{J(q)}$ is the Jacobian. In our case, the Jacobians are always computed for an end-effector that is defined in the center of the link in contact.

One of the main disadvantages of EF is that it requires an accurate dynamic model of the robot (see [18] for a discussion on the topic) or a set of empirical measurements (see [19]). However, there are some examples of using EF in terms of safety. In [20] the authors propose a path planner that ensures safety based on EF in the movement direction. In [21] an impedance controller is used to control the robot in such a way that it reduces EF during contact and thus reduces the $F_{\text{real}}$.

In Figure 3 we provide examples for our robot, where the EF in every configuration of our task is shown for three skin pads of the UR10e robot (the pads that can be in collision with any of the obstacles). We can see that EF is in most cases lower than HMM, which was the motivation to use this property in our dynamic threshold setting policy. However, we can also see that it is higher in the beginning of the task for one of the pads. The given skin part (gray pad in Figure 4) is in the beginning of the movement very close to the robot base and is moving in such a way that realistically it cannot exert any force to obstacles (see the video on https://youtu.be/UyOa--6Jpj4).

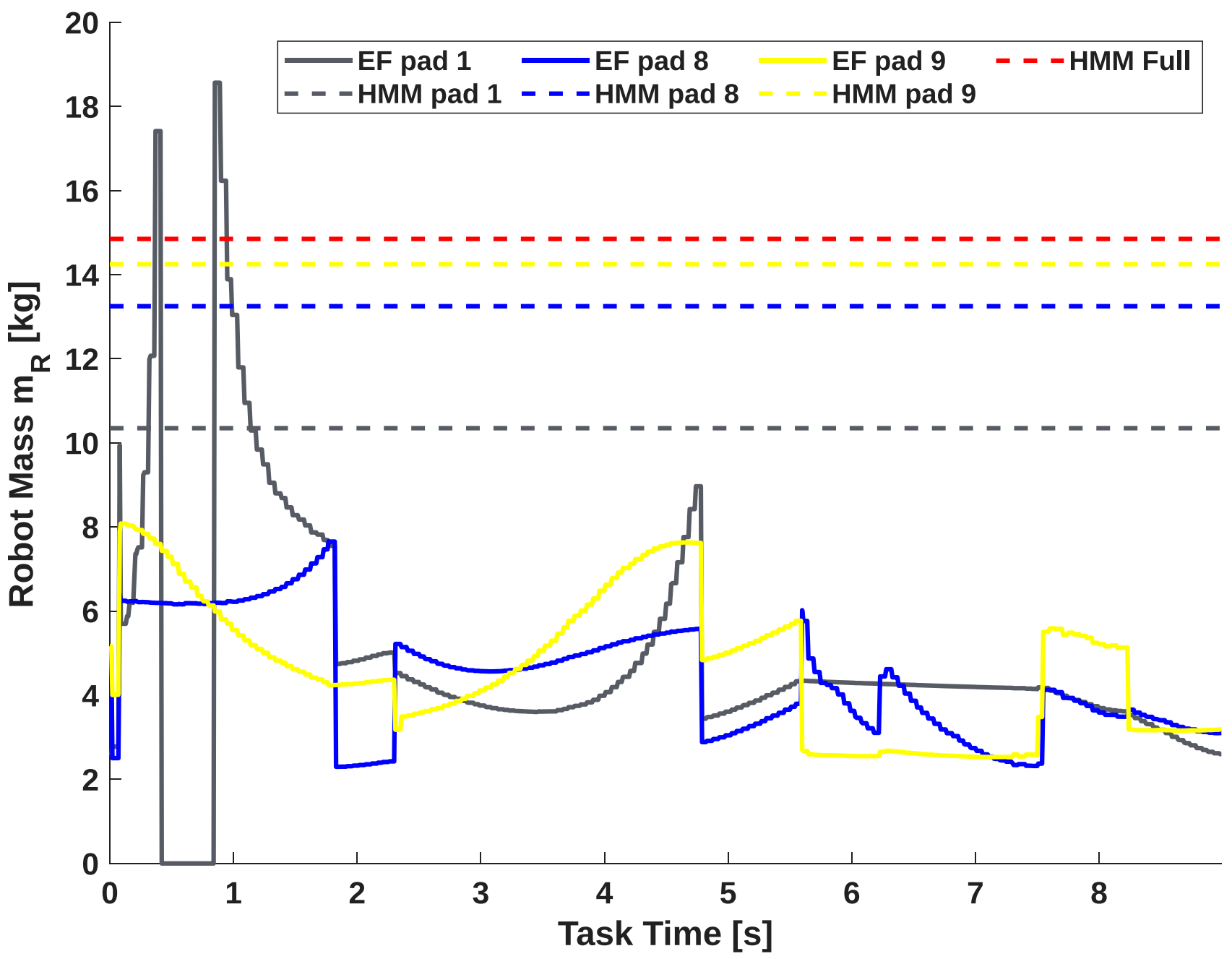

**FIGURE 3** | Effective Mass (EF) and Half of Moving Mass (HMM) for three skin parts in every configuration of the performed task for the UR10e robot. HMM Full stands for HMM of the end-effector of the robot—as in ISO 10128-2:2025 [6]. Colors correspond to colors in Figure 4.

An advantage of this numerical solution is its speed. Calculating $F_{\mathrm{est}}$ of one link takes less than 1 ms. Therefore, the total loop execution latency of the pipeline depends primarily on the underlying communication middleware and sensor polling rates. For example, deploying the UR10e with AIRSKIN yields an end-to-end control bandwidth of approximately 25 Hz, bounded by the native 30 Hz publishing rate of the skin controller. We configured the simulated KUKA joint-torque feedback loop to match this frequency for a balanced comparison.

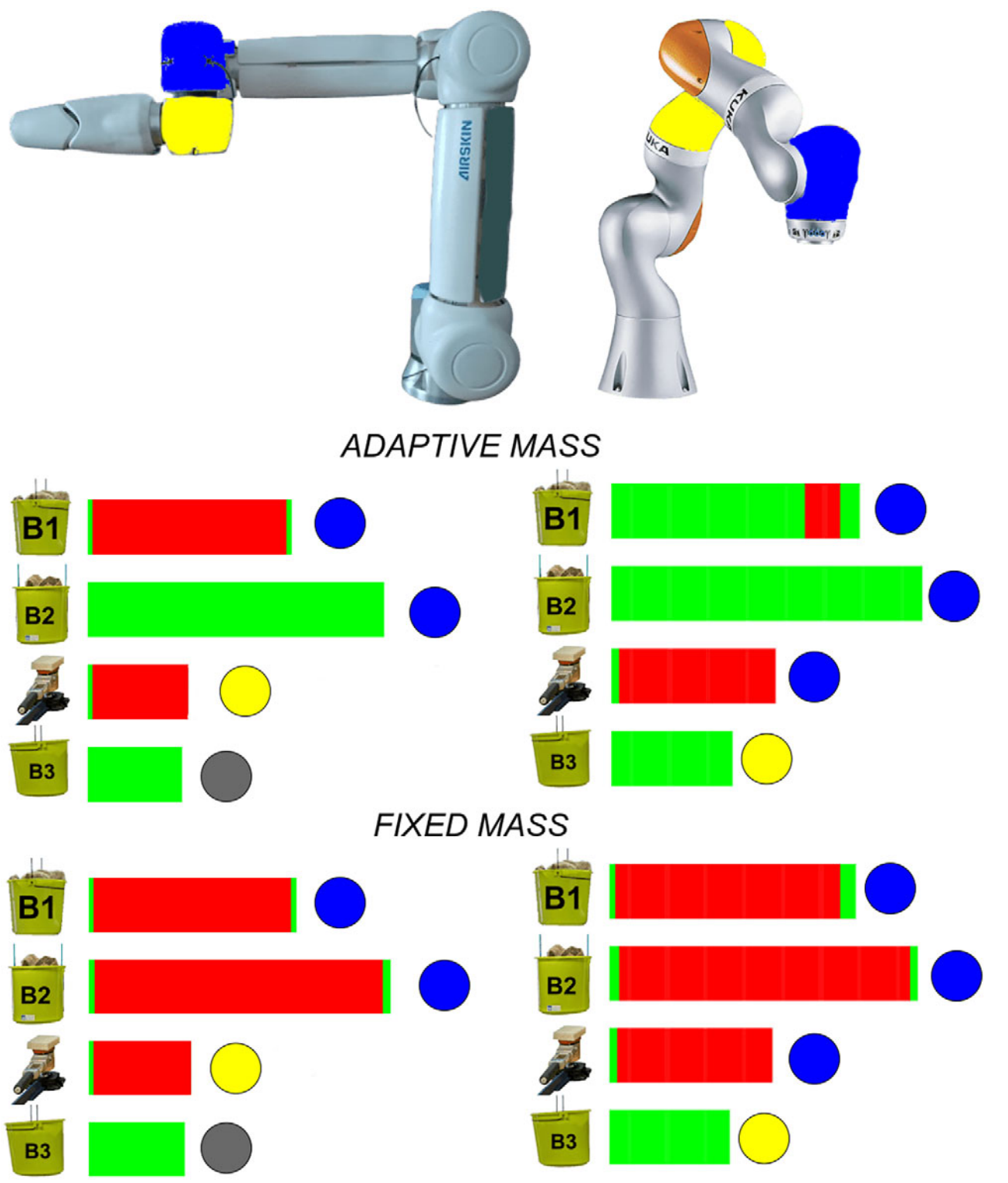

**FIGURE 4** | Schematic overview of the algorithm deployed on the UR10e robot with AIRSKIN (Left) and KUKA LBR iiwa 7 (Right). (Top) The colored parts of the robots are used for collisions in our experiments—the colors are used just for illustration and to relate robot links to collision locations in the workspace. (Bottom) The colored circles correspond to colors of the links. The color bars show the collision reaction changes over time (red = stop; green = continue) during the movements in a given axis (see Section 2.1), and end with a collision with given objects. The length of bars is proportional to the length of the movement in the given axis. The two methods (Fixed Mass [FM] and Adaptive Mass [AM]) are described in Section 2.3.

## 2.4 | Assumptions and Parameters

The proposed framework assumes several properties: (i) the robot can detect collision; (ii) there is only one collision at a time; (iii) there is a dynamic model available for the robot; (iv) it is known which obstacles result in transient and which in quasi-static contact; (v) the impact force $F_{\mathrm{est}}$ is computed in the direction $\mathbf{u}$ corresponding to the movement direction of the colliding robot link.

In addition, we use some fixed parameters shown in Table 1. The parameters are taken from ISO 10218-2:2025 [6] and correspond to the Maximum Allowed Impact Force ($F_{\mathrm{lim}}$) for the back of the hand (the commonly used worst-case scenario), while the weight of the buckets ($m_{\mathrm{H}}$) was selected as the average weight of the human arm from the same standard.

## 2.5 | Transient Contact Force Estimation

In Physical Human–Robot Interaction (pHRI), safety is at stake, and we need to make sure that the dynamic estimates of the impact forces $F_{\mathrm{est}}$, which are used to boost productivity here, are accurate or conservative (actual collision forces $F_{\mathrm{real}}$ are always lower than estimated). In our setup, the certified force-measuring device is used only at the location of the quasi-static contact. In order to empirically evaluate also the $F_{\mathrm{real}}$ of the transient collisions with the buckets suspended from the ceiling, we measured their deflection after impact and approximated them as pendulum. Therefore, a calibrated visual grid (protractor) was mounted behind bucket B1 to capture the resulting post-impact swing trajectory via video recording, allowing us to extract the maximum deflection angle $\theta$. We know that potential energy $p.e. = mgh$ and kinetic energy $k.e. = \frac{1}{2} m\omega^2 L^2$, where $m$ is the weight of the bucket, $L$ is the length of the string, $\omega$ is the angular velocity and height $h = L(1 - \cos\theta)$. Using the law of conservation of energy, the energies at the maximal angle should be equal, that is, $k.e. = p.e.$, and from there we can solve for $\omega$ as

$$\omega = \sqrt{\frac{2g}{L}(1 - \cos\theta)} \tag{4}$$

For this calculation, we used the baseline, FA, setting in which the robot stops immediately after impact. We assume that all momentum $p_{\mathrm{r}}$ will be transferred to the bucket, that is, $p_{\mathrm{r}} = p_{\mathrm{b}}$, with momentum of the bucket given by $p_{\mathrm{b}} = mv = mL\omega$. From the impulse–momentum principle we know that the average force over a small period of time $\Delta t$ is equal to the moment, that is, $F_{\mathrm{pen}}\Delta t = p_{\mathrm{r}}$. If we combine all and solve for $F_{\mathrm{pen}}$, we get

$$F_{\mathrm{pen}} = \frac{mL\sqrt{\frac{2g}{L}(1 - \cos\theta)}}{\Delta t} \tag{5}$$

**TABLE 1** | Fixed parameters used during the experiments.

| Parameter | Value | Unit |
|---|---|---|
| Mass, $m_{\mathrm{H}}$ | 5.6 | kg |
| Stiffness, $k$ | 75 000 | N/m |
| Quasi-static threshold | 140 | N |
| Transient threshold | 280 | N |

The Equation (5) holds only for a point-mass pendulum on a weightless stick in an ideal and frictionless environment. Our setup is far from these assumptions, and the force computed $F_{pen}$ with the equation is thus an underestimate of the real force $F_{real}$.

## 3 | Experiments and Results

We performed several simulated and real-world experiments using our setup that mimics possible collision scenarios in human–robot collaboration. All data collected during the experiments are available at https://rustlluk.github.io/adaptive-sensitivity. The robots and an overview showing the evolution of collision classification (red – $F_{est}$ over force threshold; green – $F_{est}$ under force threshold) can be seen in Figure 4.

### 3.1 | Experiments Settings

As described in Section 2.1, the robot performed a mock pick-and-place task in a controlled environment with three buckets (5.6 kg each) suspended from the ceiling that acted as obstacles (simulation of a collision with the human arm). Collisions were detected and isolated, and based on the decision of our algorithm, the robot was commanded to stop or to continue task operation. If stop was triggered, the robot remained stopped for 1 s. If there was no longer a collision after the waiting period, the robot continued its task. If a secondary collision event occurred at the same spatial coordinate (typically caused by the suspended bucket swinging back into the robot link), the pipeline re-evaluated the impact and triggered an appropriate independent reaction. In addition to the buckets, an impact-measuring device was attached to the table, which simulated a clamping scenario between the robot and a human operator. In case of this contact, if a stop was triggered, the robot stopped for 1 s and was commanded to go to the next waypoint of the task.

The robots were controlled in joint velocity so that the end-effector moved with a predefined Cartesian velocity. In simulation, we tested both robots with Cartesian velocities of 200, 400, and 600 mm s$^{-1}$. For each velocity and each collision strategy (FA, FM, AM) we performed 100 repetitions of the experiments. In the real world, we tested the UR10e robot with velocities of 400 and 600 mm s$^{-1}$ and performed 10 repetitions for each velocity and sensitivity setting strategy. In the real world, we had to make the movement toward the quasi-static collision at max 400 mm s$^{-1}$ even with the additional padding to not trigger the robot's internal safety controller.

### 3.2 | UR10e with AIRSKIN

We will start with the UR10e robot, for which we performed experiments in both the simulation (left side in Figure 5) and the real world (right side in Figure 5). Note that the boxplots from the simulated and physical environments should not be directly compared in absolute numbers. There is an inherent discrepancy between the two domains—primarily due to the physical pendulum dynamics and the settling times of the suspended buckets—which affects both the total task time and the absolute number of stops. Additionally, sample sizes differ, with 100 repetitions conducted in simulation versus 10 in the real world. However, the relative performance trends between the distinct collision-handling methods remain fully consistent and comparable across domains. Furthermore, the statistical differences between individual methods are evaluated using the non-parametric Wilcoxon test, which is inherently robust to unequal sample sizes and outliers. The first, and arguably most important result in both worlds is the difference in task completion time between different collision handling strategies for the Cartesian velocity of 600 mm s$^{-1}$. As expected, the FA mode, where every touch results in a stop of task execution, is the slowest. It is followed by the FM and AM modes. In the real world, the difference between the FA and AM modes is, on average, more than 6 s (45% faster; 14.5 s vs. 8 s) while FM is closer to AM with 10 s. In the simulation, the times are closer to each other with 11.5, 9, and 7.5 s, respectively. The discrepancy is probably caused by the physics of the buckets in the simulation. We tried to set the parameters as closely to the real world as possible (see the video at https://youtu.be/UyOa--6Jpj4), but the behavior is not identical.

The completion time of the task is correlated with the number of stops of the robot during the task, shown in the bottom part of the Figure 5. We can see that the trend in differences between the individual modes is the same as for the task time. In the real world, the robot stopped using AM only half the times (on average) compared to the FA mode (4 vs. 8 stops), while FM is again between them with 6 stops. In some cases, there are more stops than obstacles. This is caused by the fact that the buckets swing away from the robot and can hit it again, causing another collision detection. This behavior is supported by Figure 4, where, mainly for bucket B2, the AM strategy is more permissive. The average number of stops is lower in the simulation with 4.5, 3.5, and 2.5 for FA, FM, and AM, respectively.

With lower velocities, there is basically no difference between AM and FM, as the velocities are too slow to exceed stop thresholds for either of the strategies. In the FA mode, the task time (and the number of stops) are notably high. For example, in the real world with the velocity of 400 mm s$^{-1}$ (top right plot in Figure 5) the difference in task time is 21 s for FA versus 6.3 s for the other two, which is a speedup of over 70%. All the mentioned results are also supported by the pair-wise statistical differences between the methods—we selected the Wilcoxon test, as there are outliers in the data and we have only 10 repetitions in the real world. We can see that for the velocities (in both worlds) of 200 and 400 mm s$^{-1}$ the results for AM and FM are similar—there is no statistically significant difference, but both are significantly different from FA. For the fastest velocity, all the modes are significantly different from each other. In the real world, the $p$-values are higher, but still below 0.01.

The velocity of 400 mm s$^{-1}$ may seem slow at first glance, but if one had to follow the standard and Equation (1), then for the UR10e robot (weight of 30 kg) the maximal safe velocity is only 130 mm s$^{-1}$ for quasi-static contact and 260 mm s$^{-1}$ for transient contact. We can see that for the velocity of 400 mm s$^{-1}$ the robot stopped on average only once during the task for both modes and both worlds. Furthermore, at a velocity of 200 mm s$^{-1}$, the robot never had to stop when using neither of AM or FM. That shows that the equation from ISO 10218-2:2025 is hugely overestimating

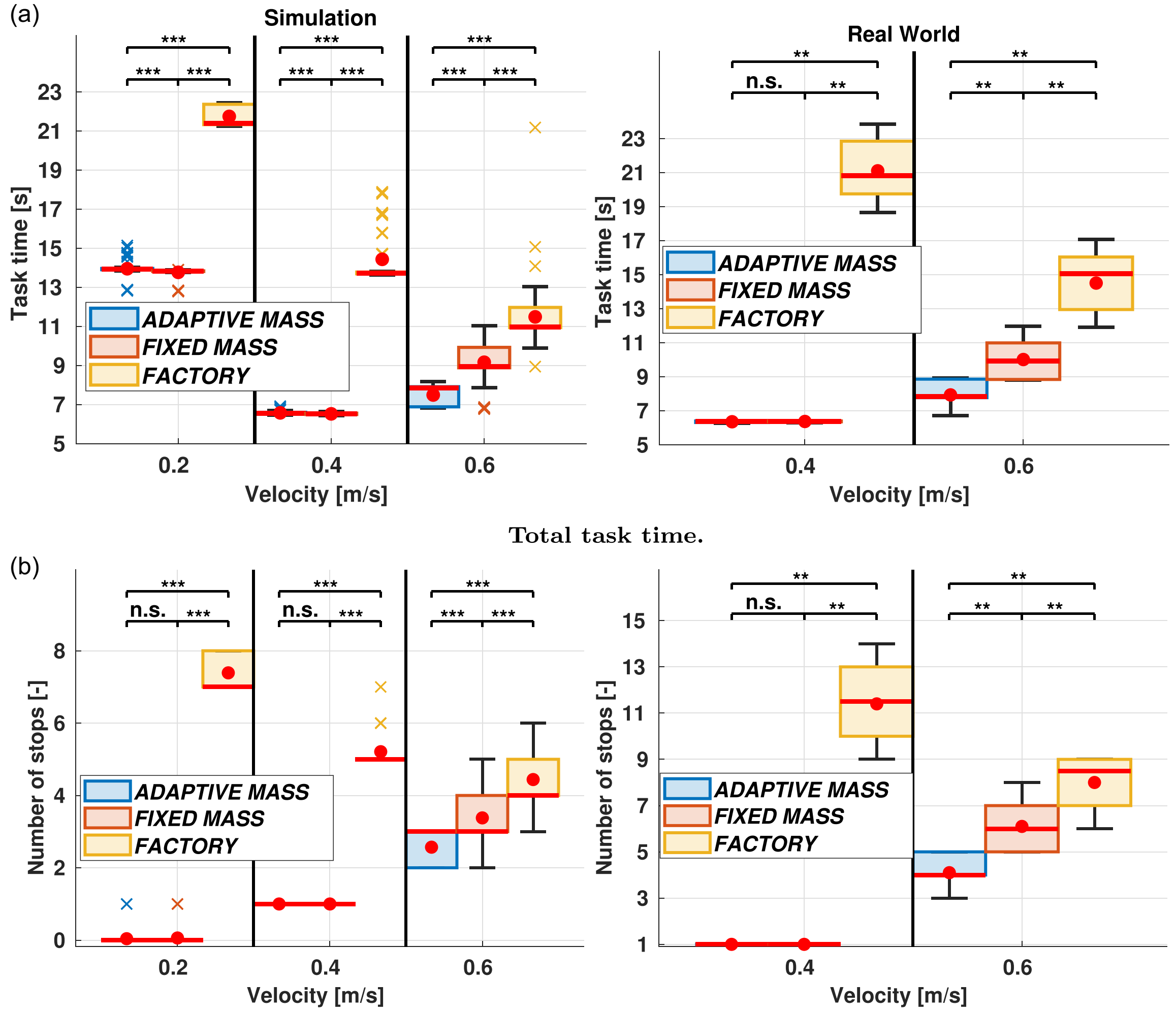

Total task time.

Number of times the robot stopped during the task.

**FIGURE 5** | Results—UR10e with AIRSKIN. Results from the simulation (left) and the real world (right). The red points represent mean and red lines median. Boxes are from 25th to 75th percentile with whiskers showing extreme non-outlier point and crosses showing outliers. Boxes are computed over 100 repetitions in the simulation and over 10 in the real world. Statistical significance was evaluated using the Wilcoxon signed-rank test, where *, **, *** denote $p$-values below 0.05, 0.01, and 0.001, respectively.

the real force $F_{real}$, and robots can actually move much faster and still not violate the force limits $F_{lim}$.

As our pipeline is composed of several components, we also measured the total latency from detected contact to total stop of the robot to assess its effect on safety. We extracted the latencies in the real world with the velocity of 600 mm s$^{-1}$ during the collision with the first bucket. The median latency for all methods was 0.22 s, resulting in the median distance traveled after contact detection of 1.5 cm. As the total latency is lower than 0.5 s (the limit for transient contact), we can consider the latency to be within safety bounds.

To further validate that our method does not compromise safety (to verify that the impact force estimates $F_{est}$ are conservative), we utilized the *CBSF-75-Basic* certified measuring device and measured the forces $F_{mes}$ exerted by the robot during quasi-static contact (the limit based on ISO 10218-2:2025 is 140 N). We measured 60 impacts in the real world, all with the velocity of 400 mm s$^{-1}$. The measuring device is sensitive to the impact "accuracy" (it should be in the middle and in the direction of the spring of the device) and to the exact time point when a stop command was executed, and thus the variance of the measurements is high. The average force $F_{mes}$ was 98 ± 44 N. As discussed in Section 2.1, without the foam we added on top of the measuring device, the maximum force $F_{real}$ would be about 20% higher (117 ± 53 N). The values estimated by AM and FM are 175 and 210 N, respectively. The estimates $F_{est}$ are higher than the real forces $F_{real}$ and higher than the safety threshold of 140 N. Therefore, both are conservative and thus safe based on the ISO standard. Our results are just a showcase of the method. In a real deployment scenario, more precise empirical measurements should be done to asses the safety at every point of the task where quasi-static impacts are expected.

As described in Section 2.5, we also attempted to measure the impact force $F_{pen}$ exerted during transient contacts using a pendulum approximation. We extracted the maximum angles after collision from the real videos of a protractor behind bucket B1. We used only values from experiments with the FA strategy, as it always stopped and it was easier to extract the angles. The measured angles were 10.3° ± 0.9° and 16.1° ± 0.9° for the velocities of 400 and 600 mm s$^{-1}$, respectively. The forces $F_{est}$ estimated using AM near bucket B1 for the velocities are 190 and 280 N, respectively. For FM these are even higher with 215 and 320, respectively. Pendulum forces $F_{pen}$ computed using Equation (5) for different contact durations $\Delta t$ are shown in Table 2. It is complicated to estimate the correct $\Delta t$, but from the recordings we know that it is more than one frame and less than two, that is, 20–40 ms. We can see in Table 2 that the estimated forces $F_{pen}$ for 40 and 50 ms are low. As both the bucket and AIRSKIN are slightly deformable, a contact duration of around 20–30 ms is a reasonable estimate. Even with the lowest estimate of $\Delta t$, the forces are more than 60 N for 400 mm s$^{-1}$ and 80 N for 600 mm s$^{-1}$ lower than the estimates from AM. As discussed, the pendulum estimation is an underestimation of the real forces $F_{real}$. But the difference is high enough to consider our method conservative.

**TABLE 2** | Average contact forces $F_{pen}$ calculated for different estimates of contact duration for pendulum approximation.

| Contact duration, ms | Force $F_{pen}$ at 400 mm/s | Force $F_{pen}$ at 600 mm/s |
|---|---|---|
| 20 | 126 ± 11N | 198 ± 10N |
| 30 | 84 ± 7N | 132 ± 7N |
| 40 | 63 ± 5N | 99 ± 5N |
| 50 | 50 ± 4N | 79 ± 4N |

### 3.3 | KUKA LBR iiwa 7

To demonstrate the general applicability of our method—it can work on any robot capable of collision detection and isolation—we also evaluated the task on a simulated KUKA LBR iiwa 7 robot. This robots does not have sensitive skin, but each of its joints is equipped with joint torque sensors that allow to detect and isolate collisions. We ran an adapted version of the same task with buckets at different places, as the robot is smaller than the UR10e.

The results are shown in Figure 6. We can see that the trends are similar to those for the UR10e. At the highest velocity (600 mm s$^{-1}$) the task time is the highest for FA mode (11 s) followed by FM (10 s) and AM (8 s), resulting in a speedup of about 30% when using AM. The number of stops again corresponds to the task time with 4.5, 4, and 3 touches, respectively. Interestingly, at lower velocities AM is slower (and stops more) than FM. It shows that FM can actually underestimate the situation in some cases and the force computed $F_{est}$ using AM is higher, which is an important aspect of safety. For this robot, the Wilcoxon test $p$-values are below 0.001 for all pairs of methods for all velocities tested.

## 4 | Conclusion, Discussion, and Future Work

We introduced three different collision handling strategies: One static and commonly used in the industry, where every detected collision triggers a stop of the robot, and two dynamic strategies that are calculating Estimated Collision Force ($F_{est}$) on the fly to choose how to react to the collisions—stop or continue operation. We tested the strategies using two robots in simulation (UR10e with artificial skin AIRSKIN and KUKA LBR iiwa 7 with joint-torque sensors) and with UR10e robot in the real world. The results show that when utilizing one of our dynamic methods (AM), which calculates the estimated force $F_{est}$ using a standardized equation from ISO 10218-2:2025 [6] and EF of the robot, the task time can be significantly lower (up to 70% with the velocity of 400 mm s$^{-1}$ or 45% with 600 mm s$^{-1}$, for the UR10e robot). The strategies also result in different number of interruptions during the task, which can be crucial for the health of the robot and its components.

The results are consistent for the two different robots we used, each of them using a different collision detection and isolation strategy (sensitive skin vs. joint torque sensing). Other collision detection methods are used by collaborative robots (a force/torque sensor in the robot base or using motor currents). Our method is potentially applicable to any of them, provided that

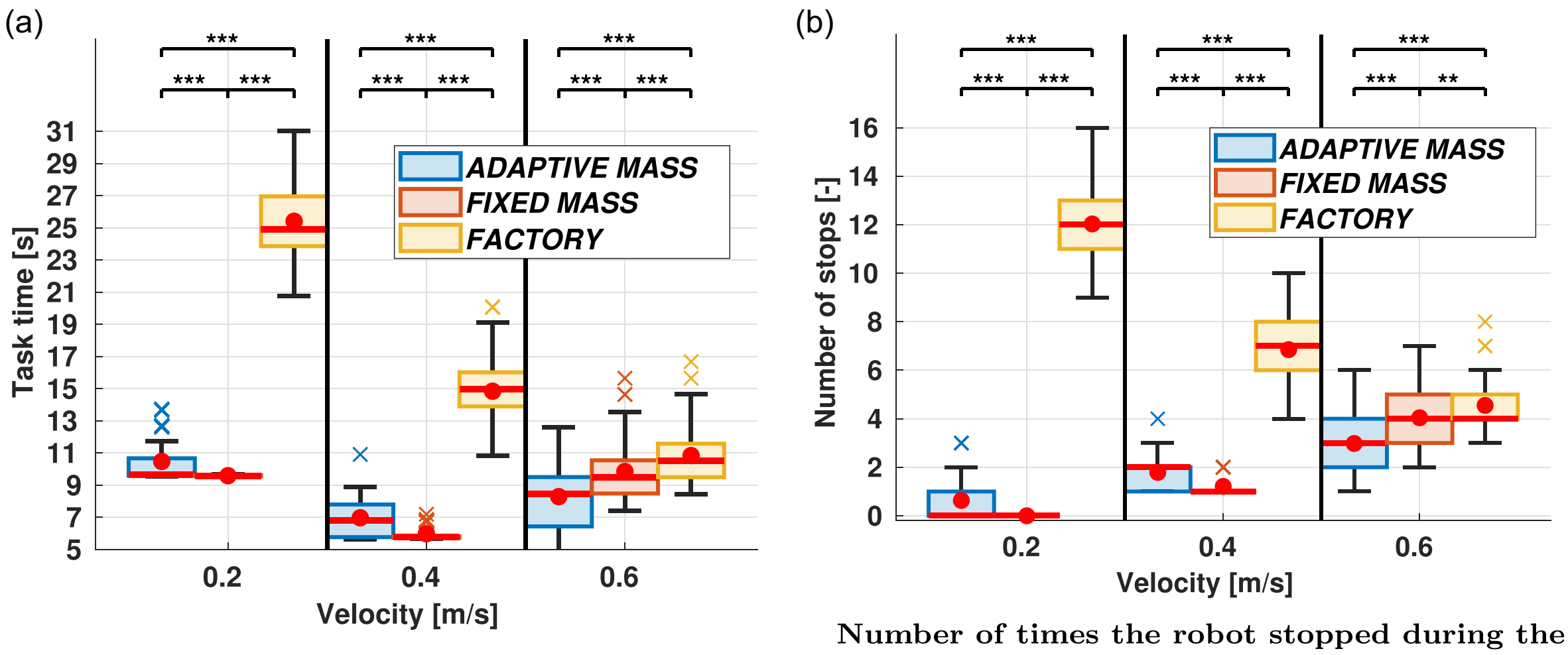

Total task time.

Number of times the robot stopped during the task.

**FIGURE 6** | Results—KUKA LBR iiwa 7 in simulation. The red points represent mean and red lines median. Boxes are from 25th to 75th percentile with whiskers showing extreme non-outlier point and crosses showing outliers. Boxes are computed over 100 repetitions. Statistical significance was evaluated using the Wilcoxon signed-rank test, where *, **, and *** denote $p$-values below 0.05, 0.01, and 0.001, respectively.

collision detection and isolation (which robot link collided) is available. Having the correct link isolated is crucial knowledge needed as we compute the EF for the given link and if the collision would be isolated on a different link, the resulting decision could not be safe (even adjacent links may have different EF and therefore different safety evaluation). Therefore, this method is applicable only if the collision isolation can be guaranteed and we would not suggest using it otherwise.

For the strategy that increases productivity the most (AM), a highly-precise dynamic model must be available to estimate EF of every robot link online. Imprecise models or unmodeled payload will alter the system's dynamics. Discrepancies between the model and reality can lead to an underestimate of EF and the resulting contact force $F_{real}$, which directly compromises the safety of the system. In real-world deployments, however, this issue can be mitigated because industrial tasks are typically repetitive, meaning that the robot's state and payload are predictable at any given time. This information can be seamlessly integrated into the online mass computation. Alternatively, for such repetitive tasks, empirical data can be collected to estimate the EF directly, bypassing the need to compute it solely from the theoretical model [19]. Using this empirical method can be beneficial, as we believe that some empirical measurements will always be indispensable before real deployment to ensure safety—see [13, 22] for an empirical comparison of safety for collaborative robots with and without whole-body contact sensing.

In our experiments, we know which contact is transient and which is quasi-static based on the collision location. It is something that is highly environment- and application-dependent and should be tailored in-situ. For example, in the case where there is no cage around the robot, it can be assumed that all collisions will be transient except for the case where a robot is moving toward the surface where it is mounted (table, ceiling, etc.). One can always also assume the worst case scenario and use the quasi-static limits. Naturally, this will make the pipeline less effective. However, it will still be more effective than FM and for a lower speeds also than FA (with higher speed the method will basically act the same as the estimated force $F_{est}$ will always be higher than the threshold).

To evaluate our framework, we measured real quasi-static impacts using a certified measuring device and estimated transient collisions using a pendulum approximation. We acknowledge the limitations in our current safety validation. In the case of quasi-static contacts, the necessary addition of foam—to prevent the robot's internal safety controller from triggering prematurely—altered the test configuration, reducing the maximum physical force $F_{real}$ (and also the measured force $F_{mes}$) by approximately 20%. However, even when this reduction is taken into account, the measured forces $F_{mes}$ remain significantly lower than our theoretical estimates $F_{est}$ (98 N measured vs. 175 N estimated for AM), indicating promising conservative behavior. For the transient pendulum estimation, our mathematical model assumes a point-mass pendulum on a weightless string in an ideal, frictionless environment. This simplification, compounded by uncertainties in estimating exact collision durations due to camera frame rate limitations, inherently underestimates the true impact forces $F_{real}$. Because our results show that the forces $F_{est}$ estimated using AM are comfortably higher than these pendulum estimates $F_{pen}$, we maintain that the method exhibits conservative tendencies. The current experimental setup demonstrates the feasibility of our approach rather than complete safety validation for real-world deployment. Rigorously proving safety in future work will require better measuring tools. For quasi-static impacts, this could involve a more compliant measuring device to prevent the robot from entering an error state. For transient contacts, this requires high-speed camera analysis with known material stiffness or testing with a standardized pendulum with precisely known physical properties, as demonstrated in [18].

In addition, we assume that the impact direction on the link where the collision was isolated is the instantaneous motion vector of that robot link. This holds true in our controlled experiments. However, in particular, for complex robot geometries, this assumption may not always hold in the real scenarios. EF for a point on a robot is mathematically modeled as an ellipsoid and can be higher in different directions than the assumed one. A worst-case scenario occurs when a robot with an angled, wedge-like cover or end-effector performs a sweeping motion and the impact occurs on this oblique surface. Although the projection of velocity magnitude is lower, the EF could still make the resulting force $F_{est}$ higher than in the assumed direction. A robust mitigation strategy would be to either estimate the surface normal at the contact point (this information is not available in our setups; sensitive skin with high spatial resolution would be needed as without a proper sensor it is hard to estimate even the contact point [23]) or evaluate the force $F_{est}$ (EF and projection of velocity magnitude) on a unit sphere around the contact point to find the true maximum, which is a straightforward change to the algorithm.

In this work, we assume that only one contact occurs at a time. This is a simplification that goes well with our controlled setup. We believe that in the expected scenarios (human robot collaboration in factories) this will also hold true in many cases (collisions should be rare and hit one limb/part of the human). However, this cannot be guaranteed. The proposed pipeline checks the status of every skin part every time new data are available. If multiple contacts are detected, the decision is taken as follows: (i) $F_{est}$ of no pad is above the threshold → no reactions are triggered; (ii) $F_{est}$ of at least one pad is above the threshold → reaction is triggered. We iterate the skin part in a fixed order (from the base to the end-effector), and thus it will always react to a first collision from the start of the kinematic link. In this work, only the simplest reaction to a collision that exceeds a force threshold was applied: stop the robot—for which the reaction to any collision works and our assumption of one contact at a time does not change neither the safety nor effectiveness of the method. Post-collision reactions can have several forms (see, for example [24]) and may also depend on what information is available about the collision (see collision identification in [9]) and on the robot capabilities (e.g., whether torque control is available [8]). In the setup used here—with coarse isolation of the collision location in space, a pressure value reading, and a velocity control mode—one possibility is to implement an avoidance reflex that will move the colliding robot link in the opposite direction (see [15]). For this, the assumption of one collision at a time is necessary; otherwise, a more complicated planner would be required that takes into account all collisions. Artificial skins with higher spatial resolution make it possible to move away the collision point along the normal to the robot skin at that location [25].

We estimate the impact force $F_{est}$ using the robot's EF. Estimating the force from the pressure readings from the skin sensor as used in, for example, ref. [15], constitutes an alternative. However, AIRSKIN pads have a non-uniform sensitivity as shown in [13] and their pressure readings can hardly serve as reliable force estimates. More accurate location and force readings could be used using a different technology (e.g., skin normal force calibration as in ref. [26]) instead of EF. However, note that EF can also be used for the estimation of impact force $F_{est}$ prior to contact.

Our method is usable for contacts with parts of the robot that are not sharp or dangerous otherwise. However, the forces and possible injuries change based on the colliding part of the robot (see [27] or [28]), which should be incorporated into the pipeline.

The method presented here operates in the PFL regime and only considers impact forces and post-collision reactions. These can be complemented by pre-collision methods where the robot can stop or slow down before contact (see [29, 30] for a combination of PFL and SSM maximizing productivity). If visual sensors are available, the actions of the human operator, with some uncertainty, may be predicted and incorporated into the robot motions, making it more legible. Additionally, cameras could also aid contact prediction and localization [31]. Integrating visual tracking would also allow the framework to move beyond the single body-part model utilized in this initial study. Using the back of a non-dominant hand is an often used scenario, as the limits for this part are strict compared to most other body parts ($k = 75\,000$ N/m; human mass $m_H$ of 5.6 kg; thresholds of 140 and 280 N) and because the hand of an operator is often the most likely body part to collide with a robot. Real human–robot collaboration exposes various regions of the body with distinct biomechanical limits. Because our underlying online force estimation is natively parameterized by human mass $m_H$ and stiffness $k$, the framework can readily scale to multi-region sensitivity. Using a vision system could allow to dynamically query the exact region of collision and adapt the parameters.

The method presented here maximizes productivity while ensuring safety in that the robot continues to operate after an impact that is within the prescribed force limits $F_{lim}$. If the force $F_{est}$ is higher, the robot stops immediately. However, since this is a post-collision reaction, it arrives late. Our method with dynamic EF would allow one to design robot trajectories where the EF at potential collision locations would never exceed the limits. This, along with empirical measurements, will be needed before the application is deployed.

Finally, in this work, we tested everything using our setup with buckets instead of humans even though we deal with safety in human–robot interactions. This allowed us to perform a proper analytical evaluation with repeatable and robust results in a safe environment. However, to do a proper safety oriented evaluation, the next iteration should incorporate experiments of the pipeline in environments with human subjects. We explicitly acknowledge the biomechanical gap between this proxy and real human contact, noting that this setup effectively serves as a conservative physical model for transient collisions. A hanging bucket filled with stones acts as a relatively rigid passive pendulum. In contrast, a real human arm features complex articulated multi-joint dynamics, soft-tissue compliance, variable muscular damping, and active reflexive recoil (see, e.g., ref. [32]). Because the passive bucket lacks both soft-tissue energy absorption and active withdrawal reflexes, it experiences harsher peak forces than an unconstrained human operator, who would naturally recoil away from the impact. Consequently, while the current setup successfully demonstrates the feasibility of the system and confirms that the forces remain bounded against a harsh proxy, an actively recoiling human arm would be expected to experience even lower contact forces $F_{real}$. However, the impact force is only one part of the equation for humans. Unexpected impact may still not be perceived well by the human operator, even if they are within the force limits. Designing robot trajectories (and reactions) that comply with not only physical but also perceived safety [33] constitutes our future work.

---

### Author Contributions

**Lukas Rustler**, **Matej Misar**, and **Matej Hoffmann:** conceptualization. **Lukas Rustler**, **Matej Misar:** coding and experiments. **Lukas Rustler**, **Matej Misar** and **Matej Hoffmann:** Writing. **Matej Hoffmann:** funding and resources.

### Funding

This work was supported by the European Union under the project Robotics and Advanced Industrial Production (reg. no. CZ.02.01.01/00/22_008/0004590). Lukas Rustler was additionally supported by the Czech Science Foundation (GAČR) under project (No. 26-22606S) and the Grant Agency of the Czech Technical University in Prague (grant No. SGS26/075/OHK3/1T/13).

### Conflicts of Interest

The authors declare no conflicts of interest.

### Data and Code Availability

The code and data are available at https://rustlluk.github.io/adaptive-sensitivity.

### Data Availability Statement

The data that support the findings of this study are openly available in Adaptive Collision Sensitivity for Efficient and Safe Human at https://osf.io/9sqt3.

### Endnotes

[1] The skin contains pressure sensors inside individual pads and we use a filter to detect collisions based on the changes of pressure.

[2] https://www.airskin.io/.